\documentclass[review]{elsarticle}
\usepackage{lmodern}
\usepackage[latin9]{inputenc}
\usepackage{geometry}
\usepackage{color}
\usepackage{array}
\usepackage{booktabs}
\usepackage{amsmath}
\usepackage{amssymb}
\usepackage{stmaryrd}
\usepackage{graphicx}
\usepackage{setspace}
\makeatletter

\providecommand{\tabularnewline}{\\}

\usepackage{lineno,hyperref}
\modulolinenumbers[5]
\usepackage{array}
\usepackage{amsbsy}
\usepackage{epstopdf}
\usepackage{colortbl} 
\definecolor{shadedRowColor}{rgb}{0.74,0.88,0.91}
\definecolor{lightgray}{gray}{0.9}
\usepackage{microtype}

\journal{XXX}

\@ifundefined{showcaptionsetup}{}{%
 \PassOptionsToPackage{caption=false}{subfig}}
\usepackage{subfig}
\makeatother

\begin{document}
\begin{frontmatter}

\title{An evaluation of randomized machine learning methods for redundant
data: Predicting short and medium-term suicide risk from administrative
records and risk assessments}

\author{Thuong Nguyen$^{b}$\footnote{Work done when Thuong was with Deakin.},
Truyen Tran$^{a*}$, Shivapratap Gopakumar$^{a}$, Dinh Phung$^{a}$,
Svetha Venkatesh$^{a}$}

\address{$^{a}$Center for Pattern Recognition and Data Analytics, Deakin
University, Australia\\
$^{b}$School of Computer Science and Information Technology, RMIT
University, Australia\\
 $^{*}$Corresponding author. E-mail address: truyen.tran@deakin.edu.au }
\begin{abstract}
Accurate prediction of suicide risk in mental health patients remains
an open problem. Existing methods including clinician judgments have
acceptable sensitivity, but yield many false positives. Exploiting
administrative data has a great potential, but the data has high dimensionality
and redundancies in the recording processes. We investigate the efficacy
of three most effective randomized machine learning techniques \textendash{}
random forests, gradient boosting machines, and deep neural nets with
dropout \textendash{} in predicting suicide risk. Using a cohort of
mental health patients from a regional Australian hospital, we compare
the predictive performance with popular traditional approaches \textendash{}
clinician judgments based on a checklist, sparse logistic regression
and decision trees. The randomized methods demonstrated robustness
against data redundancies and superior predictive performance on AUC
and F-measure.\end{abstract}
\end{frontmatter}


\global\long\def\LL{\mathcal{L}}
 \global\long\def\Data{\mathcal{D}}

\global\long\def\zb{\boldsymbol{z}}
 \global\long\def\wb{\boldsymbol{w}}
 \global\long\def\xb{\boldsymbol{x}}
 \global\long\def\fb{\boldsymbol{f}}
\global\long\def\bb{\boldsymbol{b}}

\global\long\def\bx{\boldsymbol{x}}
 \global\long\def\bX{\boldsymbol{X}}
 \global\long\def\bW{\mathbf{W}}
 \global\long\def\bH{\mathbf{H}}
 \global\long\def\bL{\mathbf{L}}
 \global\long\def\tbx{\tilde{\bx}}
 \global\long\def\by{\boldsymbol{y}}

\global\long\def\bY{\boldsymbol{Y}}
 \global\long\def\bz{\boldsymbol{z}}
 \global\long\def\bZ{\boldsymbol{Z}}
 \global\long\def\bu{\boldsymbol{u}}
 \global\long\def\bU{\boldsymbol{U}}
 \global\long\def\bv{\boldsymbol{v}}
 \global\long\def\bV{\boldsymbol{V}}
\global\long\def\vb{\boldsymbol{v}}
\global\long\def\hb{\boldsymbol{h}}
\global\long\def\rhob{\boldsymbol{\rho}}

\global\long\def\LL{\mathcal{L}}
\global\long\def\Data{\mathcal{D}}
\global\long\def\Wb{\boldsymbol{W}}

\global\long\def\likelihood{\mathcal{L}}

\global\long\def\mF{\mathcal{F}}

\global\long\def\mA{\mathcal{A}}

\global\long\def\mH{\mathcal{H}}

\global\long\def\mX{\mathcal{X}}

\global\long\def\dist{d}

\global\long\def\HX{\entro\left(X\right)}
 \global\long\def\entropyX{\HX}

\global\long\def\HY{\entro\left(Y\right)}
 \global\long\def\entropyY{\HY}

\global\long\def\HXY{\entro\left(X,Y\right)}
 \global\long\def\entropyXY{\HXY}

\global\long\def\mutualXY{\mutual\left(X;Y\right)}
 \global\long\def\mutinfoXY{W\mutualXY}

\global\long\def\given{\mid}

\global\long\def\gv{\given}

\global\long\def\goto{\rightarrow}

\global\long\def\asgoto{\stackrel{a.s.}{\longrightarrow}}

\global\long\def\pgoto{\stackrel{p}{\longrightarrow}}

\global\long\def\dgoto{\stackrel{d}{\longrightarrow}}

\global\long\def\logll{\mathcal{L}}

\global\long\def\dataset{\mathcal{D}}

\global\long\def\zb{\boldsymbol{z}}
\global\long\def\wb{\boldsymbol{w}}
\global\long\def\xb{\boldsymbol{x}}
\global\long\def\fb{\boldsymbol{f}}

\global\long\def\bzero{\vt0}

\global\long\def\bone{\mathbf{1}}

\global\long\def\bff{\vt f}

\global\long\def\bx{\boldsymbol{x}}

\global\long\def\bX{\boldsymbol{X}}

\global\long\def\bW{\mathbf{W}}

\global\long\def\bH{\mathbf{H}}

\global\long\def\bL{\mathbf{L}}

\global\long\def\tbx{\tilde{\bx}}

\global\long\def\by{\boldsymbol{y}}

\global\long\def\bY{\boldsymbol{Y}}

\global\long\def\bz{\boldsymbol{z}}

\global\long\def\bZ{\boldsymbol{Z}}

\global\long\def\bu{\boldsymbol{u}}

\global\long\def\bU{\boldsymbol{U}}

\global\long\def\bv{\boldsymbol{v}}

\global\long\def\bV{\boldsymbol{V}}

\global\long\def\bw{\vt w}

\global\long\def\balpha{\gvt\alpha}

\global\long\def\bbeta{\gvt\beta}

\global\long\def\bmu{\gvt\mu}

\global\long\def\btheta{\boldsymbol{\theta}}
\global\long\def\thetab{\boldsymbol{\theta}}

\global\long\def\blambda{\boldsymbol{\lambda}}
\global\long\def\lambdab{\boldsymbol{\lambda}}

\global\long\def\realset{\mathbb{R}}

\global\long\def\realn{\real^{n}}

\global\long\def\natset{\integerset}

\global\long\def\interger{\integerset}

\global\long\def\integerset{\mathbb{Z}}

\global\long\def\natn{\natset^{n}}

\global\long\def\rational{\mathbb{Q}}

\global\long\def\realPlusn{\mathbb{R_{+}^{n}}}

\global\long\def\comp{\complexset}
 \global\long\def\complexset{\mathbb{C}}

\global\long\def\and{\cap}

\global\long\def\compn{\comp^{n}}

\global\long\def\comb#1#2{\left({#1\atop #2}\right) }

\section{Introduction \label{sec:intro}}

Every year, about 2000 Australians die by suicide causing huge trauma
to families, friends, workplaces and communities\cite{australian2013causes}.
This death rate exceeds transport related mortality \cite{spiller2010epidemiological,ting2012trends}.
Worldwide, suicide remains one of the three leading causes of death
among age groups of 15 to 34 years \cite{bertolote2003suicide}. Studies
on the immediate precursors to suicide -- suicidal ideation, and attempts
-- reveal shocking statistics. The number of medically serious attempts
amount to more than 10 times the total number of suicide deaths \cite{kleiman2015introduction,goldsmith2002reducing}.
 For every attempt, two to three people seriously consider suicide
without attempting it \cite{nock2008cross}. This suggests that given
patient data, timely intervention between suicide ideation and attempts
can save lives. 

People frequently make contact with health services in the months
leading up to their suicide attempt \cite{tran2014risk,sakinofsky2005attendance,da2011emergency}.
A recent study revealed about 85\% of suicidal patients contacted
primary care providers months before their suicide attempt \cite{liu2012outpatient}.
In such scenarios, the crucial problem is to identify people at risk
\cite{huffman2012predictors,allen2013screening}, and prescribe intervention
strategies for preventing suicide deaths \cite{choi2012suicide}.
Current care practices involve assessing prescribed suicide risk factors
\cite{perry2012incidence,gonda2007prediction,gonda2012suicidal,haw2011living,sapyta2012evaluating}
and estimating a risk score \cite{waern2010does,fountoulakis2012development,stefansson2012suicide}.
However, the reliability and validation of suicide risk assessments
is not well understood in terms of predictive power, and remains a
controversial issue in risk management \cite{ryan2010clinical,large2012suicide}.
One of the reasons could be that many of the patient visits before
suicide attempts are not directly related to mental health problems
or self-harm \cite{luoma2002contact}. Also, a high prevalence of
coexistent physical illnesses was found in such patients \cite{qin2013hospitalization}.
Hence, for a better understanding of suicide risk, the suicide risk
factors need to be analyzed along with the patient clinical information
\cite{ryan2012suicide,tran2014risk}. 

In our previous work, we advocate a statistical risk stratification
model based on patient data from electronic medical records (EMR),
which outperformed clinical risk assessment practices \cite{tran2014risk,tran2013integrated,tran2014framework}.
Besides known risk factors for suicide, EMR patient data contains
demographic and clinical information, including patient history, disease
progression, medications. Two major issues are high dimensionality
and redundancy. Our previous work resorts to sparsity-inducing techniques
based on lasso \cite{tibshirani1996rsa}. However, lasso is linear
and has a tendency to discard useful information. More severely, it
is highly unstable under redundancy, leading to conflicting subsets
of explanatory risk factors under small data variations \cite{tran2014framework,gopakumar2014astabilizing}. 

Given the poor predictive power of risk assessment, we conjecture
that the link between historical risk factors and future suicide risk
may be nonlinear. Thus a good predictive method should be nonlinear
and insensitive to high dimensionality and redundancy. To this end,
we investigate three most effective randomized machine learning techniques
-- \emph{random forests}, \emph{gradient boosting machines}, and \emph{deep
neural nets with dropout} -- in predicting suicide risk. These methods
perform multiple random subspace sampling, and thus efficiently manage
high dimensionality and redundancy. All information is retained, there
is no discarding of potentially useful information. This property
is highly desirable since there are no well-defined risk factors that
are conclusive for predicting suicide \cite{pokorny1983prediction,goldney1987suicide}.
Our experiments are conducted on a real world hospital data set containing
$7,399$ mental health patients undergoing $16,858$ suicide risk
assessments. Prediction horizons (how far ahead the model predicts)
are 15, 30, 60, 90, 180, 360 days. 

We compare our proposed randomized methods with existing traditional
approaches to predicting suicide risk: sparse logistic regression
and decision trees. We also compare the performance of our methods
with clinicians who rely on an $18$ point checklist of predefined
risk factors. In our experiments, the randomized methods demonstrate
better predictive accuracy than clinicians and traditional methods
in identifying patients at risk on measures of AUC (area under the
ROC curve) and F1-score.

\section{Data extraction \label{sec:data-extraction}}

We use a retrospective cohort from \textcolor{black}{Barwon Mental
Health, Drugs and Alcohol Services, a regional provider in Victoria,
Australia. }Ethics approval was obtained from the Hospital and Research
Ethics Committee at Barwon Health (approval number 12/83).\textcolor{black}{{}
It is the only tertiary hospital in a catchment area with over 350,000
residents. The hospital data warehouse recorded} approximately $25K$
suicide risk assessments on $10K$ patients in the period of 2009-2012. 

We focus our study on those patients who had at least one hospital
visit and a mental condition recorded prior to a risk assessment.
This resulted in a dataset of $7,399$ patients and $16,858$ assessments.
Among patients considered, $49.3\%$ are male and $48.7\%$ are under
35 of age at the time of assessment. The main characteristics of our
study cohort are summarized in Table.~\ref{tab:Characteristics-of-cohort}

\begin{table}
\begin{centering}
\begin{tabular}{>{\raggedright}p{0.35\textwidth}>{\raggedright}p{0.5\textwidth}}
\toprule 
\textbf{Feature} & \textbf{Statistics}\tabularnewline
\midrule 
Demographics & \tabularnewline
\midrule 
\qquad{}Patients & 7,399\tabularnewline
\midrule 
\qquad{}Gender & Male: 49.3\%,~Female: 50.7~\% \tabularnewline
\midrule 
\qquad{}Age & \textless{}~21: 16.4\% ~21 -- 35: 28~\% \tabularnewline
\midrule 
\qquad{}Marital Status & Married: 20\%~Divorced/Separated: 11\%, Single:54.4\%\tabularnewline
\midrule 
\qquad{}Occupation & Unemployed/home duties: 16.7\% ~ Pensioner/Retired: 19.2\%\tabularnewline
\midrule 
\qquad{}Post code changes & in 12 months: 33.5\%~ in 12-24 months: 16.2\% ~ 24-48 months: 24.4\% \tabularnewline
\midrule 
Suicide risk assessment score & based 18 item checklist developed in Barwon health\tabularnewline
\bottomrule
\end{tabular}
\par\end{centering}

\protect\caption{Characteristics of suicide patient cohort. \label{tab:Characteristics-of-cohort}}

\end{table}

\subsection{Ground-truth of suicide risk}

Each risk assessment is considered as an evaluation point from which
we predict the future suicidal risk. We aim to predict multiple outcomes
for different time windows from the evaluation point -- 15, 30, 60,
90, 180 and 360 days. Future risk is determined based on a lookup
table of ICD-10 codes that are deemed risky by a senior psychiatrist,
as previously reported in \cite{tran2014risk}. Examples of risky
diagnostic codes are \emph{S51} (open wound of forearm) and \emph{S11}
(open wound of neck). These risk events are considered as a proxy
measure for suicide attempts, which are rare events. Further class
distributions are summarized in Table~\ref{tab:Outcome-class-distribution-1}.
\begin{table}
\begin{centering}
\begin{tabular}{lrrrr}
\hline 
 & \multicolumn{4}{c}{Horizon (day\textbf{)}}\tabularnewline
\cline{2-5} 
 & \textbf{30} & \textbf{60} & \textbf{90} & \textbf{180}\tabularnewline
\hline 
Risk (\%) & 1,243 (7.1) & 1,816 (10.3) & 2,294 (13.1) & 3,275 (18.6)\tabularnewline
Suicide (\%) & 24 (0.14) & 32 (0.18) & 41 (0.23) & 63 (0.36)\tabularnewline
\hline 
\end{tabular} 
\par\end{centering}

\protect\caption{Outcome class distribution following risk assessments.\label{tab:Outcome-class-distribution-1}}
\end{table}

\subsection{Feature extraction\label{sub:Features}}

Historical data prior to each assessment are used to extract features
(or risk factors), following the methodology in \cite{tran2014framework}.
There are two types of features: \emph{static} and \emph{temporal.}
Static features include \textbf{demographic information} such as age,
gender, spoken language, country of birth, religion, occupation, marital
status and indigenous status. Patient age is categorized into intervals.
Temporal features are those recorded as events or changing over time.
A history of 48 months was used and split into non-overlapping intervals:
{[}0-3{]}, {[}3-6{]}, {[}6-12{]}, {[}12-24{]}, {[}24-48{]}. For each
interval, events of the same type are counted and normalized. Interval-specific
features are then stacked into a long feature vector. The following
event groups are used: 
\begin{itemize}
\item \textbf{Life events}: Postcode changes are considered as events based
on the hypothetical basis that a frequent change could signify social-economic
problems.
\item \textbf{ICD-10 codes}. The EMR records contain ICD-10\footnote{International Statistical Classification of Diseases and Related Health
Problems 10th Revision, available at: http://apps.who.int/classifications/icd10/browse/2010/en} diagnostic codes. We map diagnoses into 30-element \textbf{Exlixhauser
comorbidities} \cite{elixhauser1998comorbidity}, as they are known
to be predictive of mortality/readmission risk. We also derive Mental
Health Diagnosis Groups (MHDGs) from ICD-10 codes using the mapping
table in \cite{morris-yates2000mapping}. The \textbf{MHDGs} provide
another perspective to the mental health code groups in ICD-10 hierarchy. 
\item \textbf{Suicide risk assessment}. At Barwon Health, protocol mandates
suicide risk assessments for mental health patients. Every patient
is required to be assessed at 3 intervals: at admission time, 91 days
later, and at time of discharge. This process is performed by clinicians
using ordinal assessments for 18 items covering all mental aspects
such as suicidal ideation, stressors, substance abuse, family support
and psychiatric service history. In our data, 62\% patients had one
assessment while 17\% of patients had two assessments. About 3\% of
patients had more than 10 assessments. For each assessment, we collect
statistics on risk factors and record the maximum values. An extreme
value in a risk factor, either at present or in past 3 months, is
a definite indicator for potential suicide. Thus we create an extra
subset of features with the maximum values: (i) Max of (overall ratings)
over time (ii) Sum of (max ratings over time) over 18 items (iii)
Sum of (mean ratings over time) over 18 items (iv) Mean of (sum ratings
over 18 items) over time (v) Max of (sum ratings over 18 items)
\end{itemize}
The feature vector is then fed into the classifier to predict future
suicide risk. The most challenge in dealing with the aforementioned
data is redundancy. A piece of information might be presented in multiple
feature groups, e.g. ICD-10 codes, MHDG codes or assessments. In this
study, we investigate the suitability of randomized classifiers against
this redundancy.

\section{Randomized machine learning\label{sec:randomized-methods}}

We now describe the randomized methods employed in this paper: Random
Forests (RF) \cite{breiman2001random}, Gradient Boosting Machine
(GBM) \cite{friedman2002stochastic} and Deep Neural Networks with
Dropout (DNND) \cite{srivastava2014dropout}. At present, these three
methods are considered as best performing techniques in data sciences
practice. The prediction is binary -- risk versus non-risk over multiple
periods of time.

\subsection{Random Forests}

A RF is a collection of decision trees. A decision tree makes a class
decision based on a series of tests on values of features. At each
test, a feature is selected from all features, and the splitting value
is chosen within the value range. At the terminal nodes, class decision
will be made. The result of this process is a highly interpretable
decision tree. However, decision trees are not very robust -- a slight
change in training data will lead to a vastly different tree. The
prediction variance, as a result, is high. Random forests aim at reducing
such variance by using many trees \cite{breiman2001random}. Each
tree is trained on a bootstrap resample of data. At each splitting
decision, only a small random subset of features is considered. The
final outcome is voted among trees.

A nice property of RF is that it handles high dimensionality well
-- at each decision step, only one feature is selected if it offers
the best improvement in predictive performance. Hence, important features
are repeatedly selected but unimportant features are ignored. Another
property is that redundancy is also taken care of -- at each step,
only a small subset of features is considered, thus the chance of
having redundancy is small.

\subsection{Gradient Boosting Machine}

Suppose the goal is to estimate a predictive function $F(\xb)$ which
has an additive form:
\[
F(\xb)=\sum_{t=1}^{T}\lambda_{t}h_{t}(\xb)
\]
where $h_{t}(\xb)$, known as ``weak learner'', and $\lambda_{t}>0$
is a small step size. In binary classification, the decision can be
made by checking if $F(\xb)\ge0$. We choose the following loss function:
\begin{equation}
L=\log\left(1+\exp(-yF(\xb))\right)\label{eq:log-loss}
\end{equation}
for binary output $y\in\left\{ \pm1\right\} $, which is essentially
the loss for logistic regression coupled with the nonlinear $F(\xb)$. 

GBM is a sequential method for minimizing the loss $L$ by estimating
a pair $\left\{ \lambda_{t},h_{t}(\xb)\right\} $ at a time. At each
step, the function is updated as $F_{t+1}(\xb)\leftarrow F_{t}(\xb)+\lambda_{t}h_{t}(\xb)$
. The weak learner $h_{t}(\xb)$ is estimated by approximating the
functional gradient of the loss function: $\nabla L=-y\left[1+\exp(yF(\xb))\right]^{-1}$.
Typically, $h_{t}(x)$ is learnt by regression trees, but other regression
methods such as neural networks are applicable. We implemented a randomized
variant of GBM \cite{friedman2002stochastic} in that each weak learner
is estimated on a portion $\rho\in(0,1)$ of training data. Further,
only a subset of features is used in building the weak learner.

In this paper we use regression trees for weak learner. Following
RF, each tree node split involves only a small sub-subset of features.
Thus this retains the capacity of handling high dimensional and redundant
data of the RF while offering more flexibility in controlling overfitting
through learning rate $\lambda_{t}$.

\subsection{Deep Neural Networks with Dropout and Multitask Learning}

Deep Neural Networks (DNNs) are multilayer perceptrons with more than
one hidden layer. We aim at estimating the predictive function $F(\xb)$
of the following recursive form:
\begin{equation}
F(\xb)=b+\wb^{\top}\sigma\left(\bb_{L}+W_{L}\sigma(...)\right)\label{eq:risk-func-DNN}
\end{equation}
where $\sigma$ is a nonlinear transformation, also known as activation
function. In the past, typically $\sigma$ is a sigmoid or tanh function,
but more recently, a rectified linear ($\mbox{\ensuremath{\sigma(x)=\max(0,x)}}$)
is used due to the ease of passing gradient in back-propagation. Here
we use the same loss as in Eq.~(\ref{eq:log-loss}). 

With sufficient non-linear hidden layers, DNNs are able to learn any
complex function $F(\xb)$ \cite{srivastava2015training}. This flexibility,
however, makes them susceptible to overfitting \cite{futoma2015comparison}.
Traditionally, parameter shrinkage methods, also known as weight decay,
are used to prevent overfitting. However, these methods do not create
an ensemble, which has been proven to be highly successful in the
case of RFs and GBMs. Second, they are not designed for high dimensionality
and redundancy.

\paragraph{Dropout}

We use a recently introduced elegant solution -- ``dropout'' \cite{srivastava2014dropout}
-- with these desirable properties. At each training step, some hidden
units and features are randomly removed. In effect, exponentially
many networks are trained in parallel sharing the same set of weights.
At test time, all the networks are averaged by weight, and thus creating
a single consensus network of the original size. The result is that
dropout achieves model averaging similar to RF but without storing
multiple networks. The use of random feature subsets also helps combat
against high dimensionality and redundancy, similar to RF and GBM.
Due to its effectiveness, dropout is considered as one of the best
advances in neural networks in the past decade. A more detailed account
of dropout is presented in ~\ref{sub:Appendix---Dropout}.

\begin{figure}
\centering{}\includegraphics[width=0.9\textwidth]{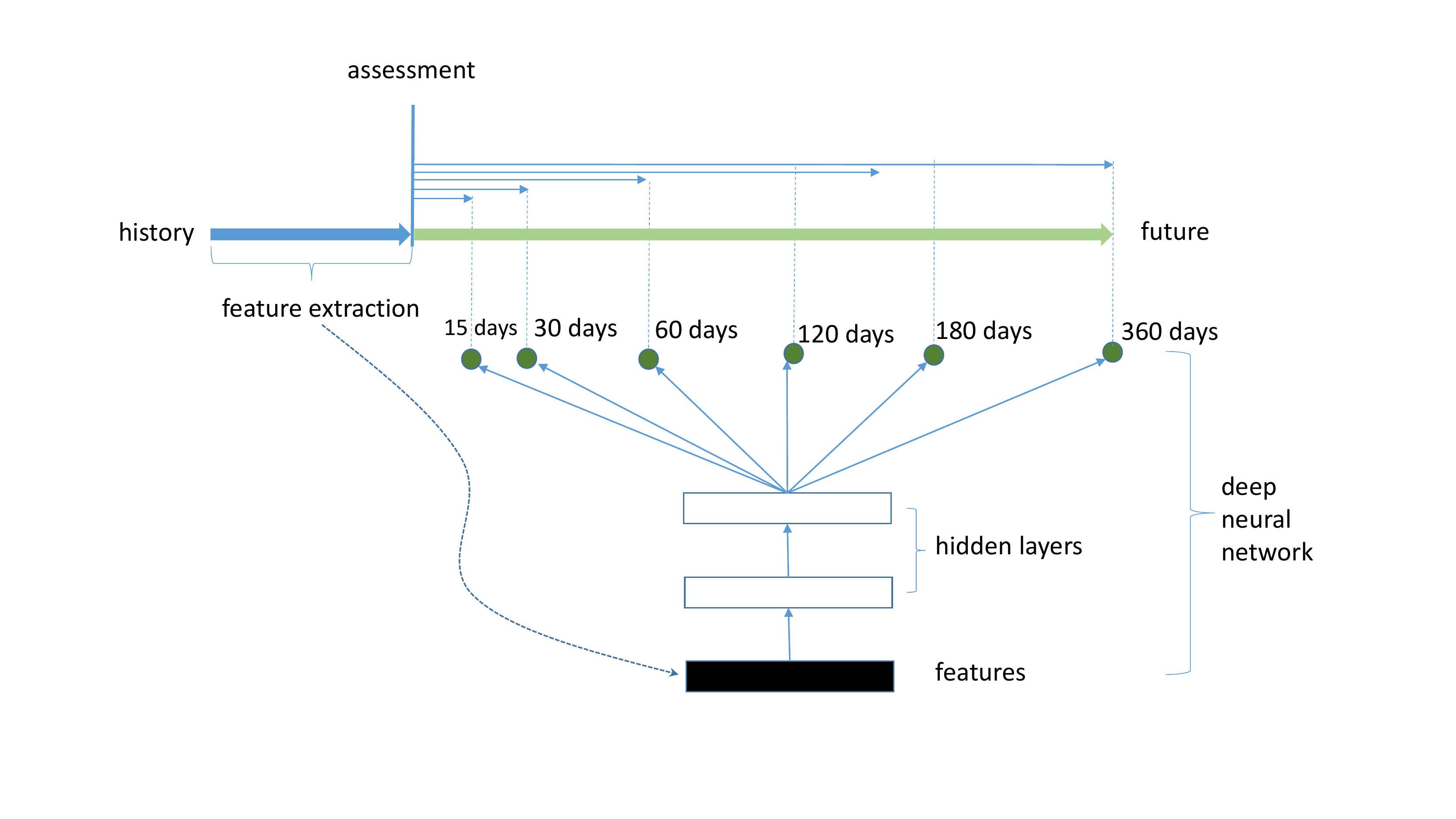}\protect\caption{Short and medium-terms suicide risk prediction using multitask deep
neural networks.\label{fig:Risk-prediction-DNN}}
\end{figure}

\paragraph{Multitask learning}

Since we are predicting risk for multiple future periods, the problem
can be considered in the multitask learning framework. Neural networks
are natural candidate as multiple outcomes can be predicted as the
same time. Eq.~(\ref{eq:risk-func-DNN}) can be extended as follows:
\[
F_{m}(\xb)=b_{m}+\wb_{m}^{\top}\sigma\left(\bb_{L}+W_{L}\sigma(...)\right)
\]
where $m$ denotes the $m$-th outcome. That is, all the layers except
for the top remain the same. The loss function is now a composite
function: $L=\sum_{m}\log\left(1+\exp(-y_{m}F_{m}(\xb))\right)$.
Learning using back-propagation and dropout is carried out as usual.
See Fig.~\ref{fig:Risk-prediction-DNN} for an illustration.

\section{Experimental results \label{sec:Experiments}}

\subsection{Experimental setup}

Here we describe our experimental setup, which is summarized in Fig.~\ref{fig:Our-experimental-setup}.

\paragraph{Feature sets}

We examine three different combinations of the features mentioned
in Section.~\ref{sub:Features}: 
\begin{itemize}
\item \textbf{Feature set \#1 (FS1, without assessment): Demographics, ICD-10
and MHDGs:} Our first feature set consists of features commonly available
in most hospital setting. It includes three groups: demographics,
ICD-10 and MHDGs. There are total 415 features from these three groups.
We filter out the features that are active for less than 1\% of data
points resulting in 109 features.
\begin{figure}
\begin{centering}
\includegraphics[width=0.8\textwidth]{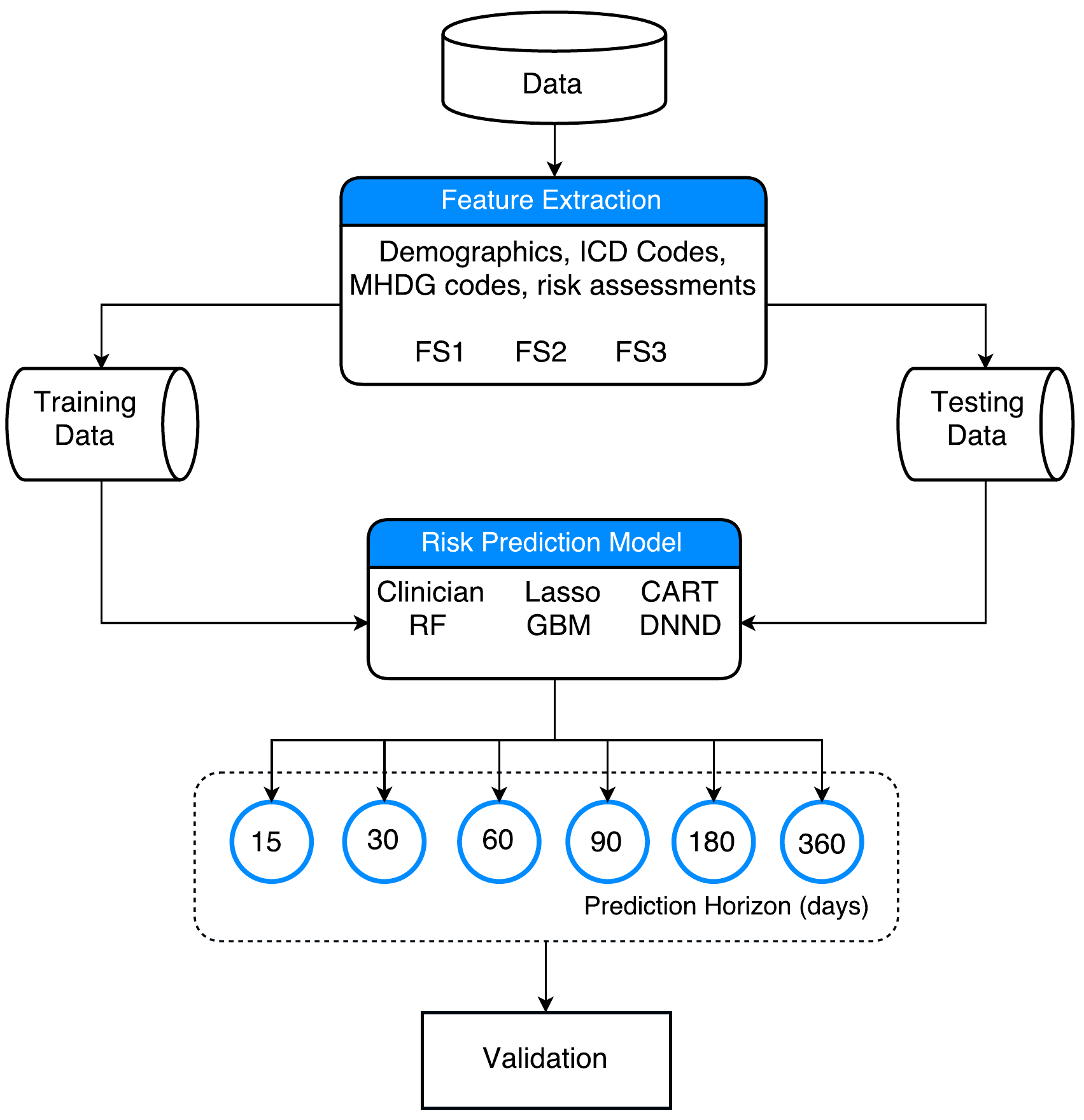}
\par\end{centering}

\protect\caption{Our experimental setup. \label{fig:Our-experimental-setup}}
\end{figure}

\item \textbf{Feature set \#2 (FS2, with assessment): Demographics, ICD
codes, MHDGs and assessments}: In the second setting, we use all available
features to form feature set \#2. These features include demographics,
ICD-10, MHDGs and assessments. This feature set differs from the feature
set \#1 in the assessments. There are total 440 features. We filter
out the features that are active for less than 1\% of data points
resulting in 134 features.
\item \textbf{Feature set \#3 (FS3, mental health information only): MHDGs
and assessments:} In the third setting, we use only two groups of
mental health features (MHDGs and assessments) to form feature set
\#3. This feature set includes 85 features in total. We filter out
the features that are active for less than 1\% of data points resulting
in 37 features. 
\end{itemize}

\paragraph{Data splitting}

The patients are randomly split into a training set of 3,700 patients
(8,466 assessments) and a validation set of 3,699 patients (8,392
assessments).

\paragraph{Baseline prediction models}

We compare the randomized methods described in Section~\ref{sec:randomized-methods}
against baselines. Three baseline techniques are: clinician assessments,
lasso regularized logistic regression (lasso-LR) \cite{friedman2010regularization},
and CART \cite{hastie2009esl}. Clinician assessment produces is an
overall score of risk based on the 18-item checklist (see also Sec.~\ref{sub:Features}).
CART generates interpretable decision trees \cite{hastie2009esl}.
Logistic regression enjoys wide popularity in medical statistics due
to its simplicity and interpretability \cite{hosmer2004applied}.
Though simple, it has proven to be very effective in many studies
\cite{austin2010logistic}, and has been used to investigate suicide
in many recent studies \cite{brent2015familial,bernert2014association}.
We use lasso regularized logistic regression to find a compact subset
of features from that best represents suicide risk \cite{friedman2010regularization}.
Lasso has one tuning parameter -- the penalty factor, which is tuned
to obtain the best performance. 

Details of the experimental setup for Random Forests, Gradient Boosting
Machine and Deep Neural Network with Dropout are presented in ~\ref{sub:Model-experiment-settings}.

\paragraph{Validation}

We consider the suicide risk prediction as a binary classification:
\emph{risk} versus \emph{non-risk}. Each assessment for a patient
is treated as a data point to predict future suicide risk. Each model
is used to predict the suicide risk at six different horizons: (i)
15 days (ii) 30 days (iii) 60 days (iv) 90 days (v) 180 days and (vi)
360 days. The classification performance of each model is evaluated
using (a) Recall $R$ (a.k.a. sensitivity), (b) Precision $P$ (a.k.a.
positive predictive value or PPV), (c) F-measure, computed as $2RP/(R+P)$,
which is a balance between recall $R$ and precision $P$, and (d)
area under the ROC curve (AUC, a.k.a. $c$-statistic) with confidence
intervals based on Mann-Whitney statistic \cite{birnbaum1956use}.

\subsection{Results}

We test 5 machine learning methods with 3 different feature sets.
The training set and validation set are split as discussed in the
experimental setup section. We feed the training set to each method
and obtain the learned models. We then use these models to predict
the output on validation set to compute recall, precision, F-measure
and the Area under ROC curve (AUC).

\subsubsection{Feature set \#1: Demographics, ICD-10 and MHDGs}

\begin{figure}
\subfloat[Recall]{\protect\includegraphics[width=0.5\columnwidth]{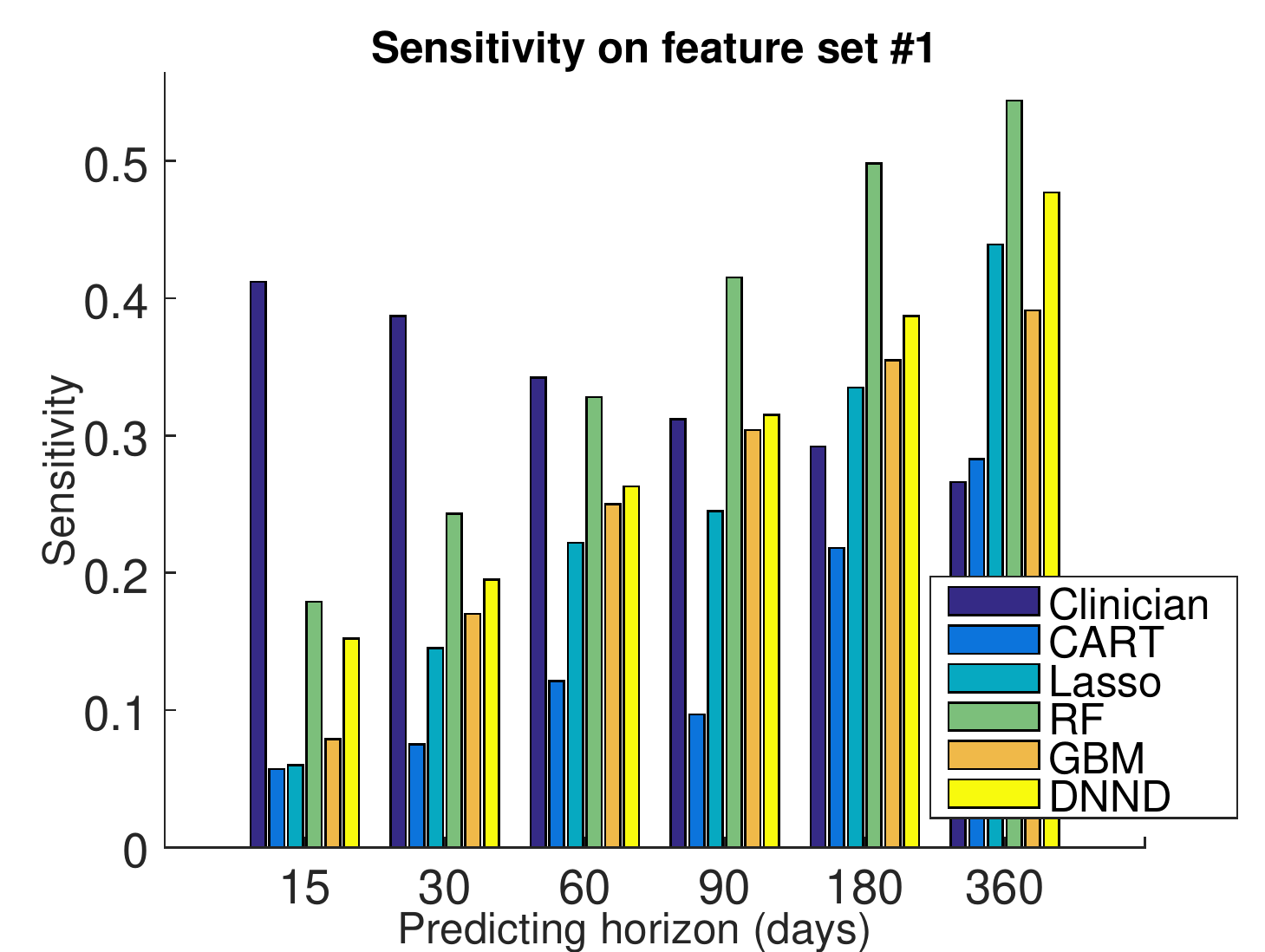}

}~~\subfloat[Precision]{\protect\includegraphics[width=0.5\columnwidth]{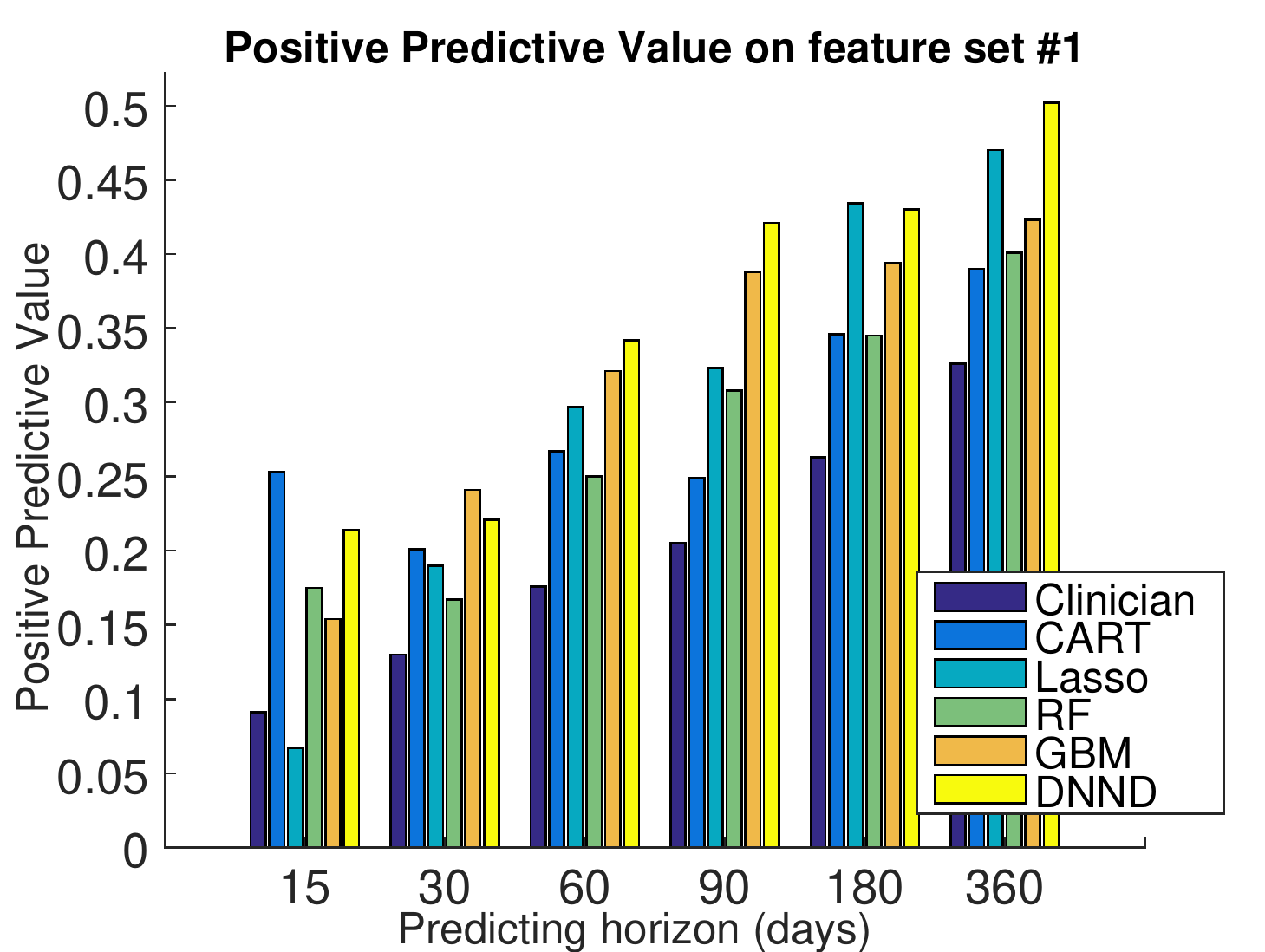}

}

\protect\caption{Recall and precision on Feature set \#1: Demographics, ICD-10 and
MHDGs.\label{fig:Recall-and-precision-FS1}}
\end{figure}

Recall and precision of all 6 methods are presented in Figs\@.~\ref{fig:Recall-and-precision-FS1}(a,b).
Clinician assessments tend to detect more short-term risk within short
terms (high recall/sensitivity) at the cost of low precision. Machine
learning methods, on the other hand, tend to be more conservative
and strike the balance between recall and precision. This is reflected
on F-measures reported in Table~\ref{tab:Fscore_featureset1}. On
this measure, CART performs poorly compared to prediction of clinician
and other methods. Its F-measure is lower than that of clinician prediction
at almost all horizons (except at 360-days horizon). Lasso-LR performs
better than clinician at mid-term horizons (60-360 days) but short-term
horizons (15-30 days). On the other hand, the randomized methods (RF,
GBM and DNND) performs better than the remaining methods and clinician,
except for the GBM at 15-days horizon. Out of these three methods,
DNND always gives the highest F-measure at all horizons and the margin
compared to lasso-LR is significant. 

\begin{figure}
\begin{centering}
\includegraphics[width=0.6\columnwidth]{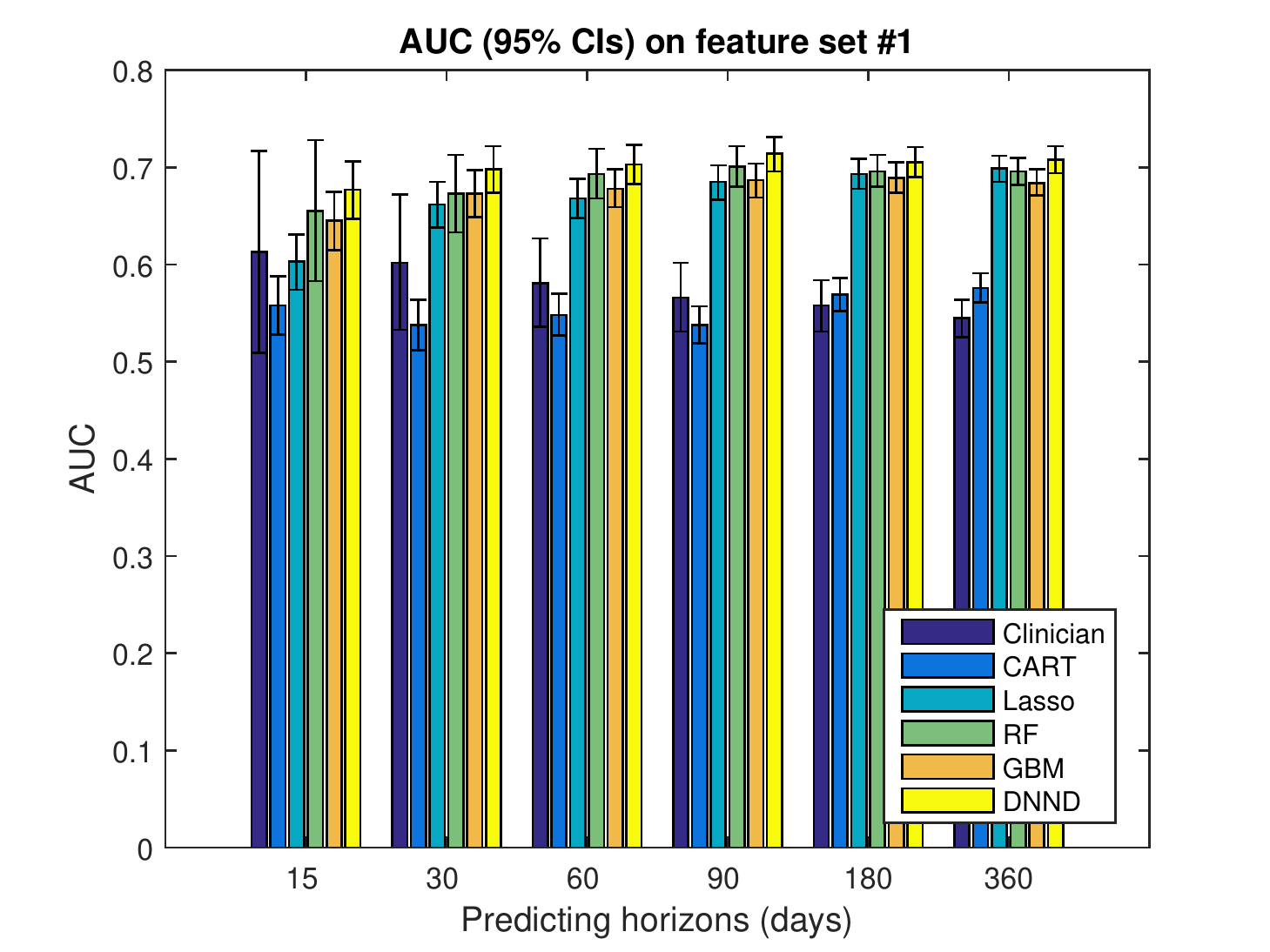}
\par\end{centering}

\protect\caption{Comparison of AUC (95\% CIs) from feature set \#1: Demographics, ICD-10
and MHDGs.\label{fig:AUC_featureset1}}
\end{figure}

A comparison of AUC obtained from this feature set over multiple predicting
horizons is presented in Figure~\ref{fig:AUC_featureset1}. Except
CART, all predictive methods outperforms clinician with significant
margins (from 6\% for 15-days horizon to 25\% for 360-days horizon).
Among predictive methods, the randomized methods always perform the
best. 
\begin{table}
\begin{centering}
\begin{tabular}{|c|c|c|c|c|c|c|}
\hline 
Horizon & Clinician & CART & Lasso-LR & RF & GBM & DNND\tabularnewline
\hline 
15 days & 0.149 & 0.093 & 0.063 & \textbf{0.177} & 0.104 & \textbf{0.177}\tabularnewline
30 days & 0.195 & 0.109 & 0.164 & 0.198 & 0.199 & \textbf{0.207}\tabularnewline
60 days & 0.232 & 0.167 & 0.254 & 0.284 & 0.281 & \textbf{0.297}\tabularnewline
90 days & 0.248 & 0.139 & 0.279 & 0.354 & 0.341 & \textbf{0.360}\tabularnewline
180 days & 0.277 & 0.267 & 0.378 & \textbf{0.407} & 0.374 & \textbf{0.407}\tabularnewline
360 days & 0.293 & 0.328 & 0.454 & 0.462 & 0.406 & \textbf{0.489}\tabularnewline
\hline 
\end{tabular}
\par\end{centering}

\protect\caption{Comparison of F-measure obtained from Feature set \#1: Demographics,
ICD-10 and MHDGs.\label{tab:Fscore_featureset1}}
\end{table}

\subsubsection{Feature set \#2: Demographics, ICD codes, MHDGs and assessments}

\begin{figure}
\subfloat[Recall]{\protect\includegraphics[width=0.5\columnwidth]{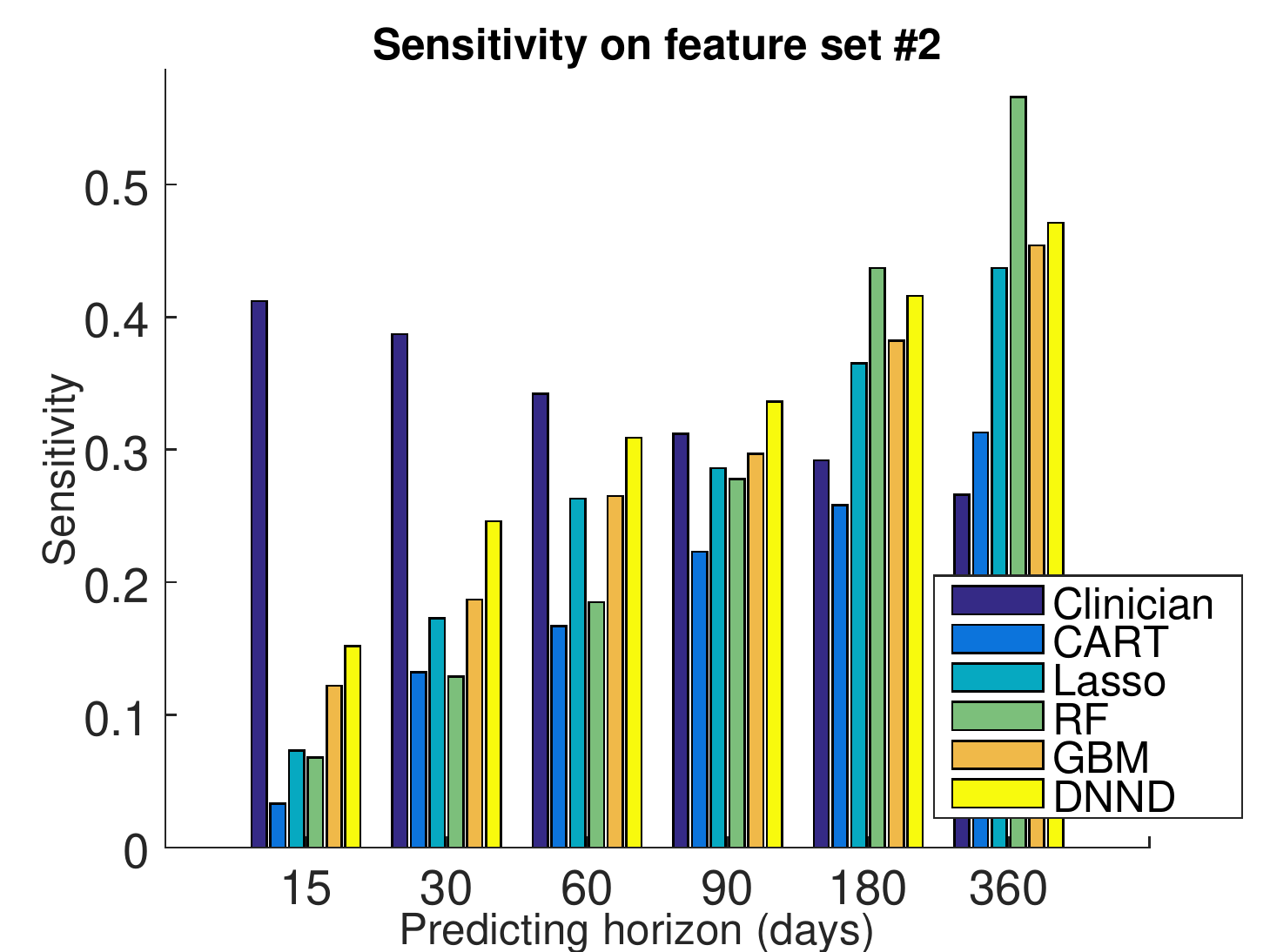}

}~~\subfloat[Precision]{\protect\includegraphics[width=0.5\columnwidth]{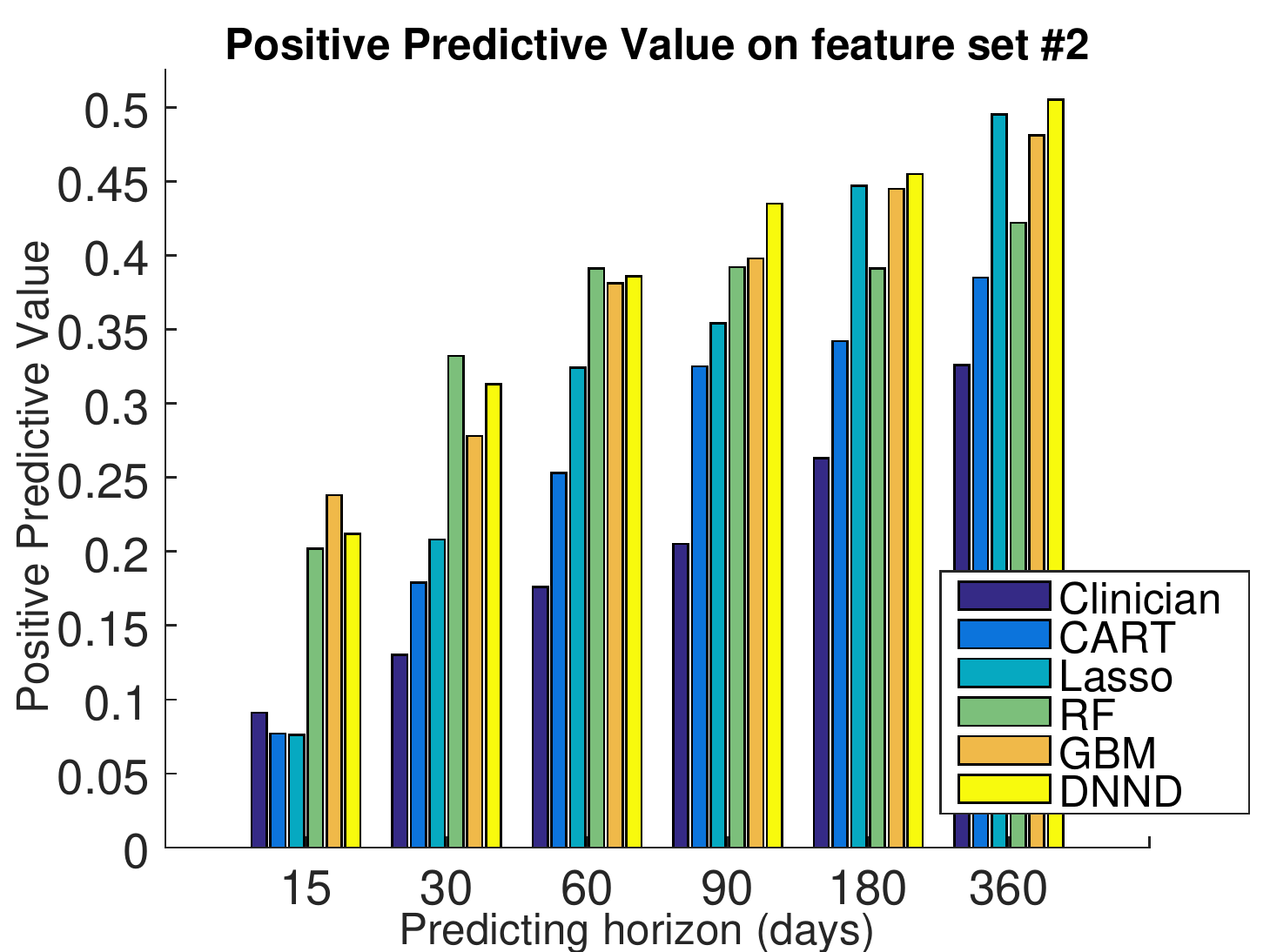}

}

\protect\caption{Recall and precision on Feature set \#2: Demographics, ICD-10, MHDGs
and assessments.\label{fig:Recall-and-precision-FS2}}
\end{figure}

\begin{figure}
\begin{centering}
\includegraphics[width=0.5\textwidth,height=0.5\textwidth]{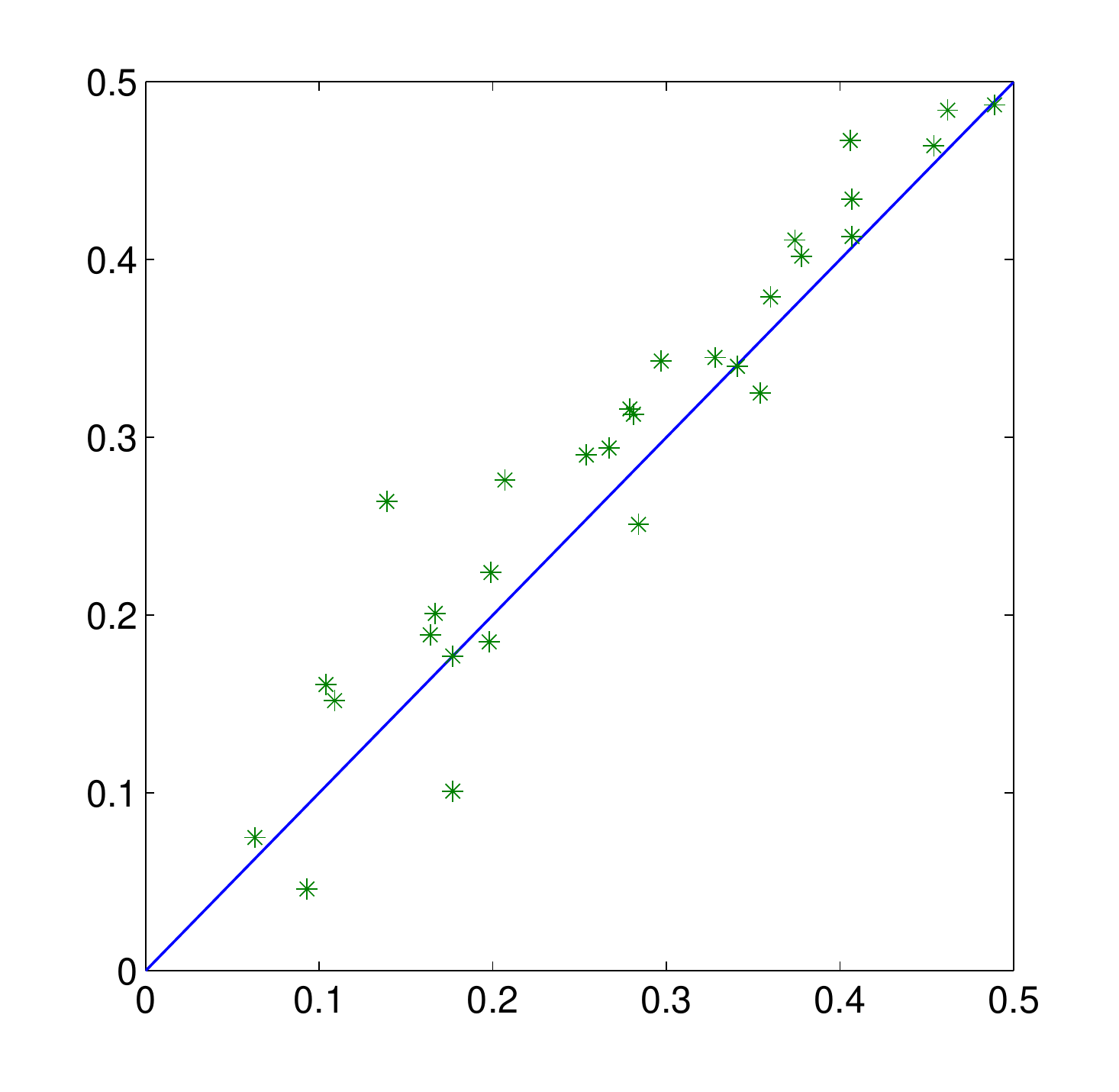}
\par\end{centering}

\protect\caption{Values of adding assessments on F-measures over all methods and all
predictive horizons. Points above the diagonal indicate improvement
due to assessments.\label{fig:Values-of-adding-assessments}}
\end{figure}

In this experiment, we investigate whether adding assessments would
improve the predictive performance. Figs.~\ref{fig:Recall-and-precision-FS2}(a,b)
show recall and precision. Overall, the results look qualitatively
similar to those found earlier using just clinical information. More
quantitatively, Fig.~\ref{fig:Values-of-adding-assessments} plots
the F-measures of feature set \#2 against F-measures of feature set
\#1 for all machine learning methods and all predictive horizons.
There are 22 out of 30 cases where adding assessments improve the
F-measure indicating that assessments may hold extra risk information
that is not readily available in the medical records. However, the
mean difference in F-measures due to assessment is merely 0.02, suggesting
that the extra risk information is not very critical. 

\begin{figure}
\begin{centering}
\includegraphics[width=0.6\columnwidth]{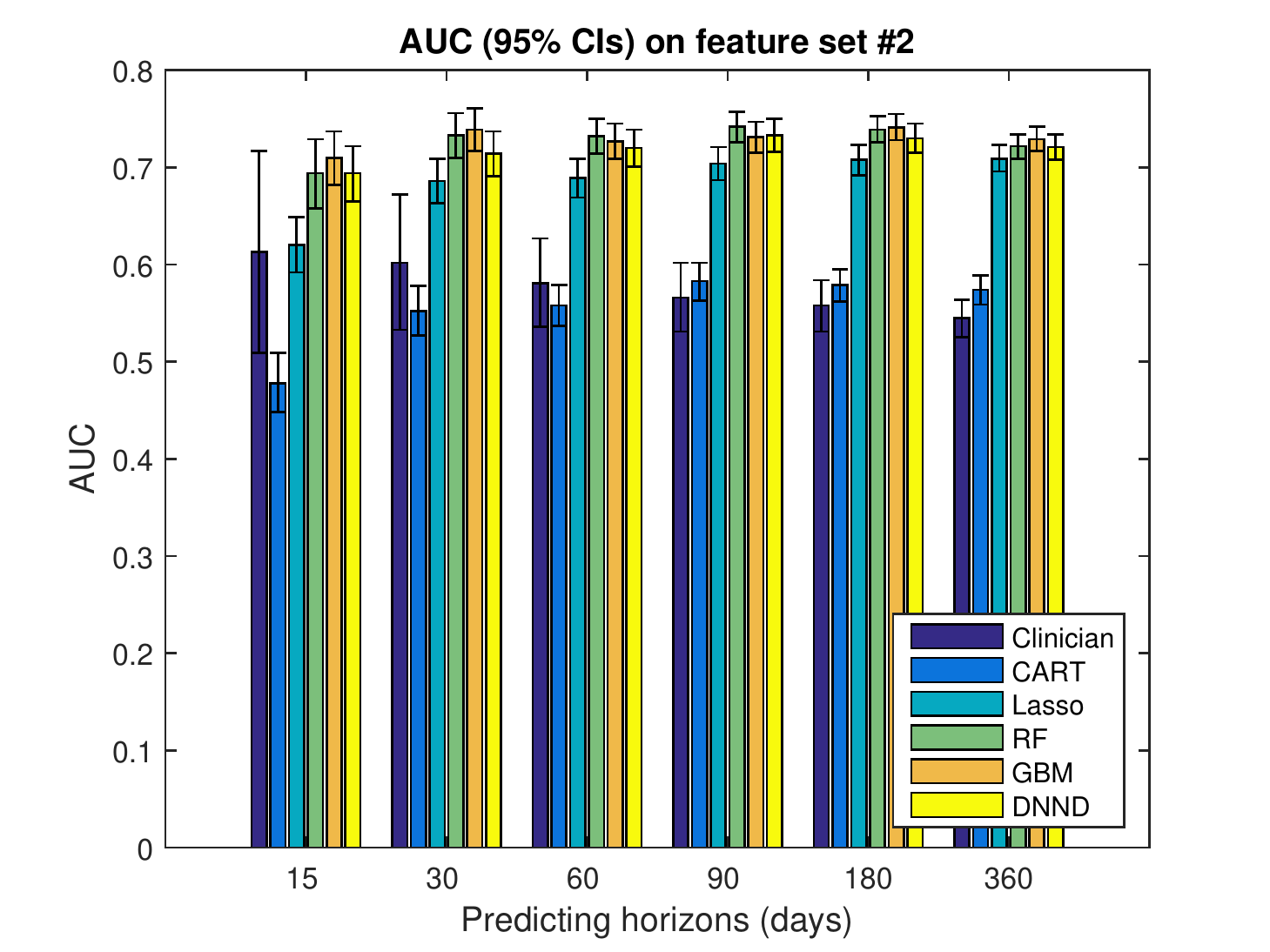}
\par\end{centering}

\protect\caption{Comparison of AUC (95\% CIs) from feature set \#2: Demographics, ICD-10,
MHDGs and assessments.\label{fig:AUC_featureset2}}
\end{figure}

Table~\ref{tab:Fscore_featureset2} reports the F-measures in detail.
DNND is still the best predictive method on this feature set. A comparison
of AUC obtained on feature set \#2 is plotted in Figure~\ref{fig:AUC_featureset2}.
Overall, AUC figures increase compared to those of feature set \#1.
Especially, AUCs obtained by 3 randomized methods are greater than
70\% (from 71\% for 15-days horizon to the highest of 74\%. These
methods outperform lasso-LR at short-term and mid-term horizons.
\begin{table}
\begin{centering}
\begin{tabular}{|c|c|c|c|c|c|c|}
\hline 
Horizon & Clinician & CART & Lasso-LR & RF & GBM & DNND\tabularnewline
\hline 
15 days & 0.149 & 0.046 & 0.075 & 0.101 & 0.161 & \textbf{0.177}\tabularnewline
30 days & 0.195 & 0.152 & 0.189 & 0.185 & 0.224 & \textbf{0.276}\tabularnewline
60 days & 0.232 & 0.201 & 0.290 & 0.251 & 0.313 & \textbf{0.343}\tabularnewline
90 days & 0.248 & 0.264 & 0.316 & 0.325 & 0.340 & \textbf{0.379}\tabularnewline
180 days & 0.277 & 0.294 & 0.402 & 0.413 & 0.411 & \textbf{0.434}\tabularnewline
360 days & 0.293 & 0.345 & 0.464 & 0.484 & 0.467 & \textbf{0.487}\tabularnewline
\hline 
\end{tabular}
\par\end{centering}

\protect\caption{Comparison of F-measure obtained from feature set \#2: Demographics,
ICD-10, MHDGs and assessments.\label{tab:Fscore_featureset2}}
\end{table}

\subsubsection{Feature set \#3: MHDGs and assessments}

\begin{figure}
\subfloat[Recall]{\protect\includegraphics[width=0.5\columnwidth]{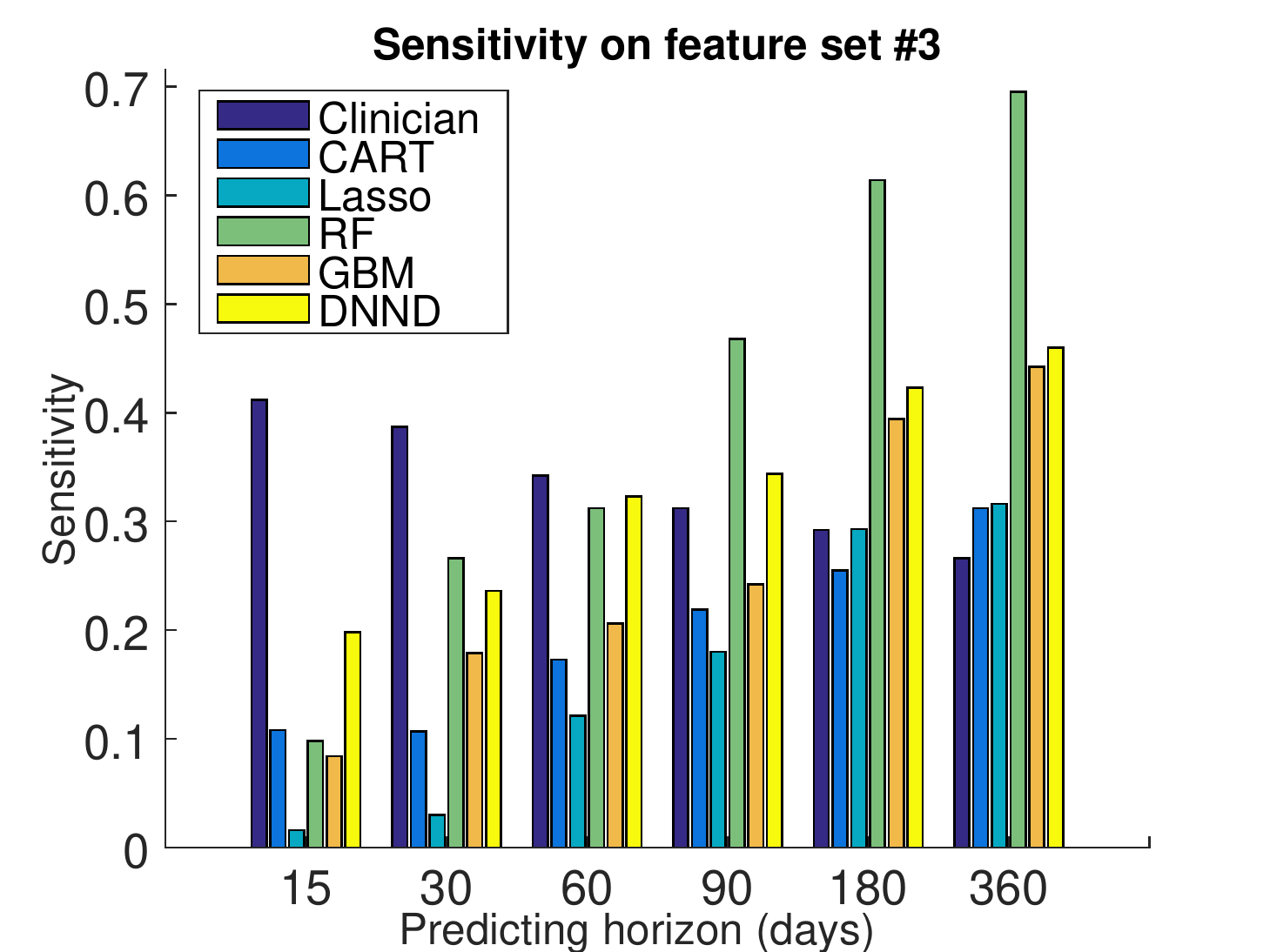}

}~~\subfloat[Precision]{\protect\includegraphics[width=0.5\columnwidth]{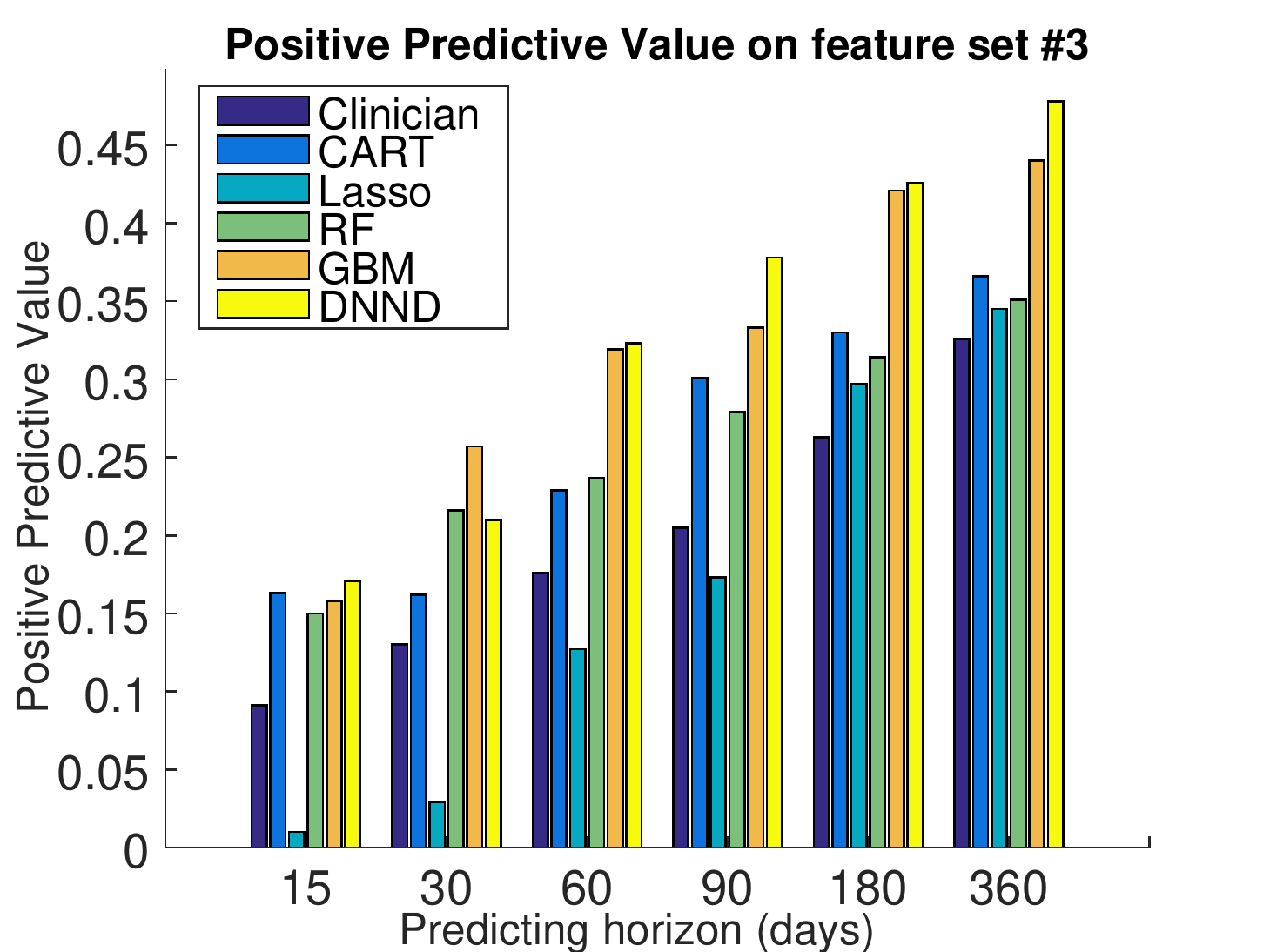}

}

\protect\caption{Recall and precision on feature set \#3: MHDGs and assessments.\label{fig:Recall-and-precision-FS3}}
\end{figure}

Recall and precision are reported in Fig.~\ref{fig:Recall-and-precision-FS3}(a,b).
A comparison of F-measure obtained on feature set \#3 is presented
in Table~\ref{tab:Fscore_featureset3}. Leaving out two groups of
features (demographics and ICD-10), F-measure metrics drop by a little
amount. However, DNND is still the best predictor, as previous two
settings.

\begin{figure}
\begin{centering}
\includegraphics[width=0.6\columnwidth]{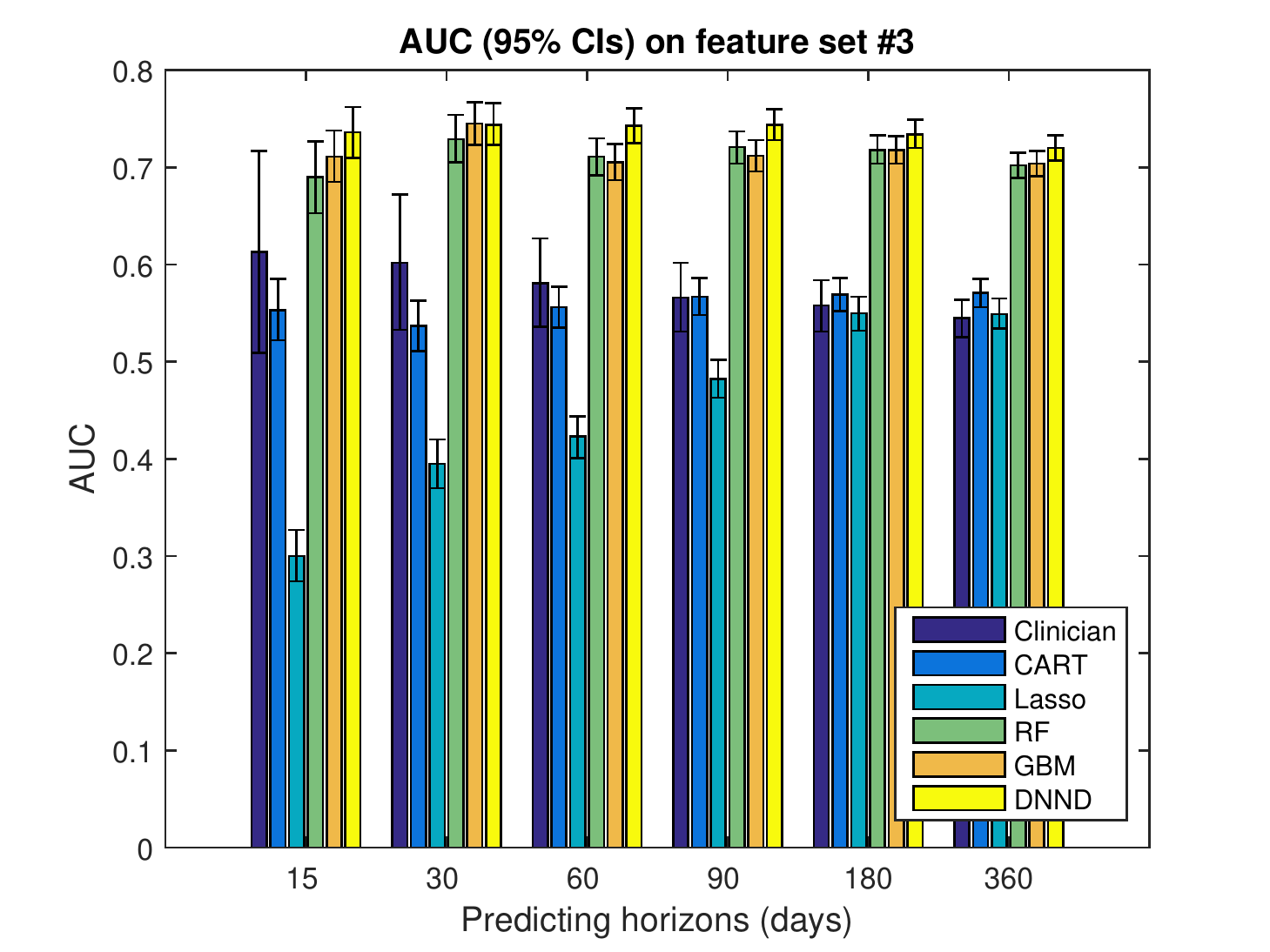}
\par\end{centering}

\protect\caption{Comparison of AUC (95\% CIs) from feature set \#3: MHDGs and assessments.\label{fig:AUC_featureset3}}
\end{figure}

A comparison of AUC obtained on feature set \#3 is plotted in Figure~\ref{fig:AUC_featureset3}.
On this feature set, AUCs of three randomized methods increase by
a significant amount on short-term and mid-term horizons. For 15-days
horizon, the highest AUC is of DNND (0.736, CIs: {[}0.710, 0.762{]}).
AUCs other short-term and mid-term horizons are greater than 74\%.
On the other hand, AUCs obtained by lasso-LR on this feature set drop
significantly, ranges from 30\% to 55\%.

\begin{table}
\begin{centering}
\begin{tabular}{|c|c|c|c|c|c|c|}
\hline 
Horizon & Clinician & CART & Lasso-LR & RF & GBM & DNND\tabularnewline
\hline 
15 days & 0.149 & 0.130 & 0.013 & 0.118 & 0.110 & \textbf{0.183}\tabularnewline
30 days & 0.195 & 0.129 & 0.030 & 0.239 & 0.211 & \textbf{0.222}\tabularnewline
60 days & 0.232 & 0.197 & 0.124 & 0.269 & 0.250 & \textbf{0.323}\tabularnewline
90 days & 0.248 & 0.254 & 0.177 & 0.350 & 0.281 & \textbf{0.360}\tabularnewline
180 days & 0.277 & 0.288 & 0.295 & 0.416 & 0.407 & \textbf{0.425}\tabularnewline
360 days & 0.293 & 0.337 & 0.330 & 0.466 & 0.441 & \textbf{0.469}\tabularnewline
\hline 
\end{tabular}
\par\end{centering}

\protect\caption{Comparison of F-measure obtained from feature set \#3: MHDGs and assessments.\label{tab:Fscore_featureset3}}
\end{table}

\section{Discussion \label{sec:Discussion}}

Predicting suicide is extremely challenging due to the rarity of the
event and absence of reliable and consistent risk factors. Ensemble
learning and model averaging combines several weak learners and improves
prediction accuracy. In this paper, we attempt to improve accuracy
of suicide risk prediction by using randomized machine learning techniques,
and compare their performance with traditional methods and clinician
assessments.

\paragraph{Findings}

In terms of predictive power (measured by F-measure and AUC), predictive
machine learning methods outperform clinician prediction. This resembles
findings in previous work using linear lasso-based methods \cite{tran2014risk}.
The new finding is that randomized machine learning methods (RF, GBM
and DNND) outperformed linear models over feature sets studied. Among
the three feature sets used to build the model, demographics and ICD-10
features had significant impact on lasso-penalized logistic regressions,
while the randomized methods only needed MHDG and assessments to make
good predictions. This could be explained by the linearity of logistic
regression, which tends to work better when more features are available
to separate the classes. Nonlinear methods can exploit the data structure
better to find nonlinear regions that correspond to risky outcomes.

While it is widely acknowledged that the final clinician rating in
risk assessments has limited predictive power \cite{ryan2010clinical}
and is highly variable among clinicians \cite{regehr2015suicide},
we found that \emph{the knowledge generated by the assessment process}
is rich, provided that there exist powerful learning methods to exploit
it. This also suggests that combining multiple assessment instruments
may offer improved accuracy \cite{blasco2012combining}.

High dimensionality and redundancies are major issues in medical records
that have led to feature selection and sparsity-inducing techniques.
Our results demonstrate that randomized methods are, by design, robust
against these properties.

\paragraph{Suicide risk prediction}

This work contributes to the literature of suicide prediction and
prevention. At present the understanding of risk factors and how they
interact is rather poor. Improving the situation is a major goal in
``A Prioritized Research Agenda for Suicide Prevention: An Action
Plan to Save Lives'', 2014 by National Action Alliance for Suicide
Prevention\footnote{actionallianceforsuicideprevention.org}. Most
existing work, however, is focused on identifying individual risk
factors. The suicide risk factors are, however, rather weak. Individual
identification will generally over-estimate the power of each factor.
Further, these factors have complicated interactions with patient
characteristics causing their predictive power to be distributed over
many aspects of patient health. 

Earlier studies focused on using statistical techniques to select
a small subset of risk factors based on their predictive power \cite{pokorny1983prediction,goldney1985suicide,goldney1987suicide}.
These methods however returned a huge number of false positives. Again,
this can be attributed to the low prevalence of suicide. A later study
using multivariate analysis of 21 common predictors failed to identify
patients who committed suicide \cite{goldstein1991prediction}. A
recent study of predicting deliberate self harm (DSH) was able to
detect high risk patients using clinical decision rules \cite{bilen2013can}.
However, in the absence of data for the specific rules, the study
performed poorly.

\paragraph{Deep learning}

Among randomized methods, we found that Deep Neural Networks with
dropout and multitask learning work best. Deep Neural Networks with
dropout have been recently shown to work well for 30-readmission prediction
\cite{futoma2015comparison}. It suggests that with recent advances,
deep learning has a great potential to play a leading role in biomedical
settings \cite{lecun2015deep}. Deep learning has multiple desirable
properties that fit biomedical data well. First, features can be learnt,
rather than designed by hand. Second, features can be learnt for multiple
tasks, as demonstrated in this paper. This can be readily extended
to multiple cohorts or transferring between domains (sites and cohorts).
Second, multiple modalities and views (such as EMR, clinical text
and medical imaging) can be integrated easily at multiple levels of
abstraction rather than at the feature levels. Third, structured data
such as temporal dynamics of disease progression or spatial imaging
can be modeled using existing techniques such as Recurrent Neural
Networks and Convolutional Networks.

\paragraph{Limitations}

We acknowledge the following limitations in our work. We used only
a single retrospective cohort and confined to a single location for
our experiments. The use of future ICD codes as proxy of suicide risk
is based on experience not internationally recognized. The use of
randomized methods is critical to obtain higher predictive accuracy
than standard logistic regression, but they are harder to tune and
interpret. However, it is possible to derive feature importance from
Random Forests, Gradient Boosting Machine and Deep Neural Networks,
and thus enables quantification of risk factor contribution.

\section{Conclusion}

As demonstrated in the experiments, randomized methods significantly
improve predictive accuracy over traditional methods. Hence they provide
valuable information to clinicians in eliminating false positives
and focusing care and resources for high risk patients. It is therefore
advisable that randomized techniques to be used for complex data and
nonlinear relationships. Concurring with \cite{futoma2015comparison},
we believe that deep learning techniques are likely to play a greater
role in the coming years in biomedical settings.

Data from EMR has been successfully used to identify suicidal patients
with high risk \cite{de2013validation,tran2014risk}. The models described
in our work are derived from routinely collected EMR data. Such models
can be easily generalized to sites with similar EMR systems. The models
based on EMR could be updated in real-time, and make use of data that
are routinely collected. The predictors derived from the EMR data
were standardised, and thus the tools can be generalizable to sites
with similar EMR systems.

\appendix

\section{Appendix }

\subsection{Details of dropout \label{sub:Appendix---Dropout}}

\begin{figure}
\begin{centering}
\includegraphics[width=0.7\textwidth]{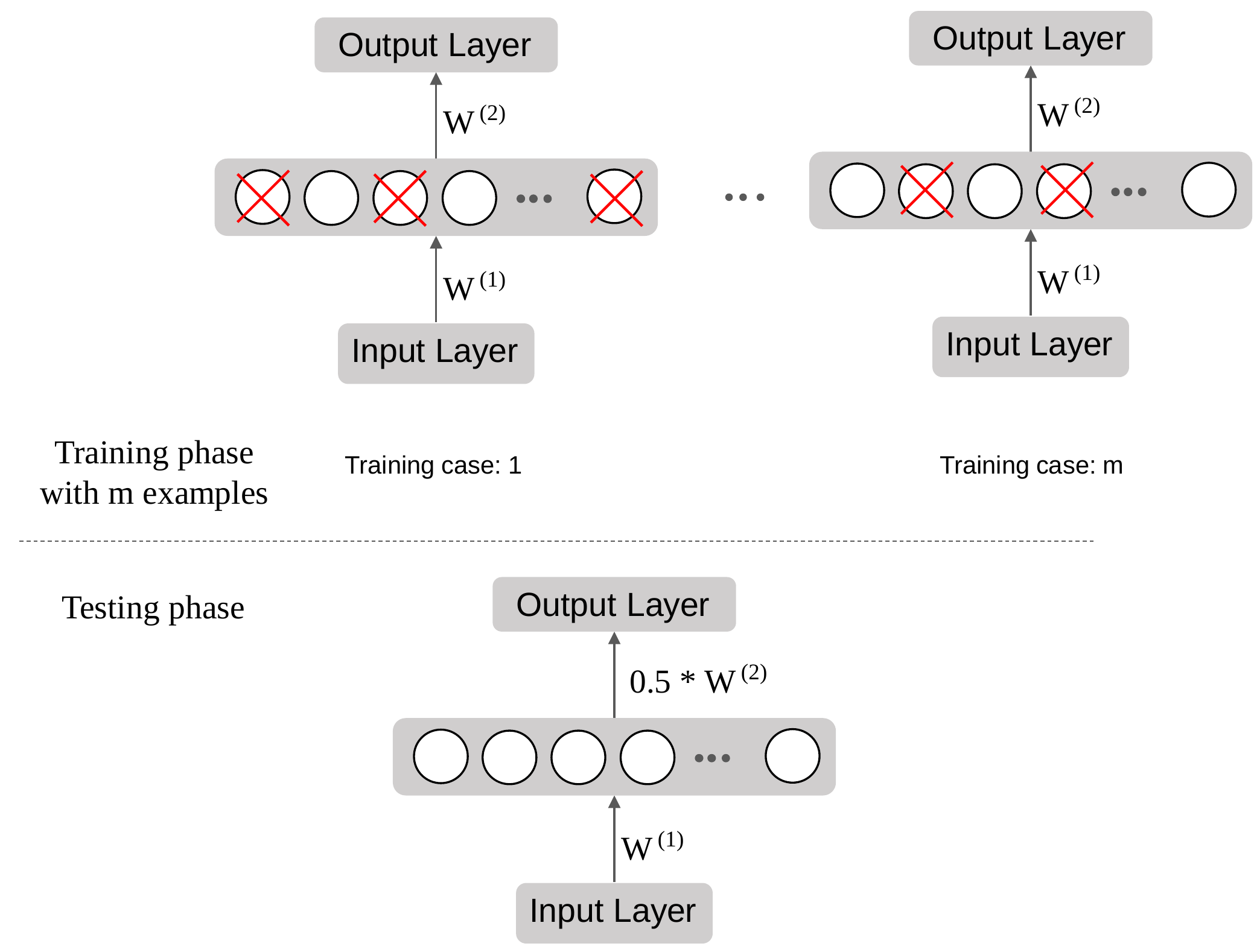}
\par\end{centering}

\protect\caption{Illustration of dropout in a single layer neural network with dropout
rate = $0.5$ \label{fig:Illustration-of-dropout}}
\end{figure}

Consider a simple scenario of a neural network with one hidden layer
containing $K$ units. For $m$ training examples, the dropout training
and testing of the network is illustrated in Fig~\ref{fig:Illustration-of-dropout}.
For every training example at a updating step, we randomly drop (or
disconnect) each hidden unit with a probability $r_{1}=0.5$. Hence,
every example trains a different network model. This is equivalent
to randomly sampling from $2^{K}$ possible models. At the end of
the training phase, many of the $2^{K}$ models will be trained from
a single training example. The weights of the hidden units are shared
among models, making it an extreme form of bagging.

The testing phase requires to average these $2^{K}$ models. An alternative
is to use all the hidden units and multiply their weights by the dropout
rate $r_{1}=0.5$ (Fig.~\ref{fig:Illustration-of-dropout}). For
a neural network with a single hidden layer and a logistic regression
output, this exactly computes the geometric mean of $2^{K}$ model
predictions \cite{srivastava2014dropout}. 
\begin{figure}
\begin{centering}
\includegraphics[width=0.7\textwidth]{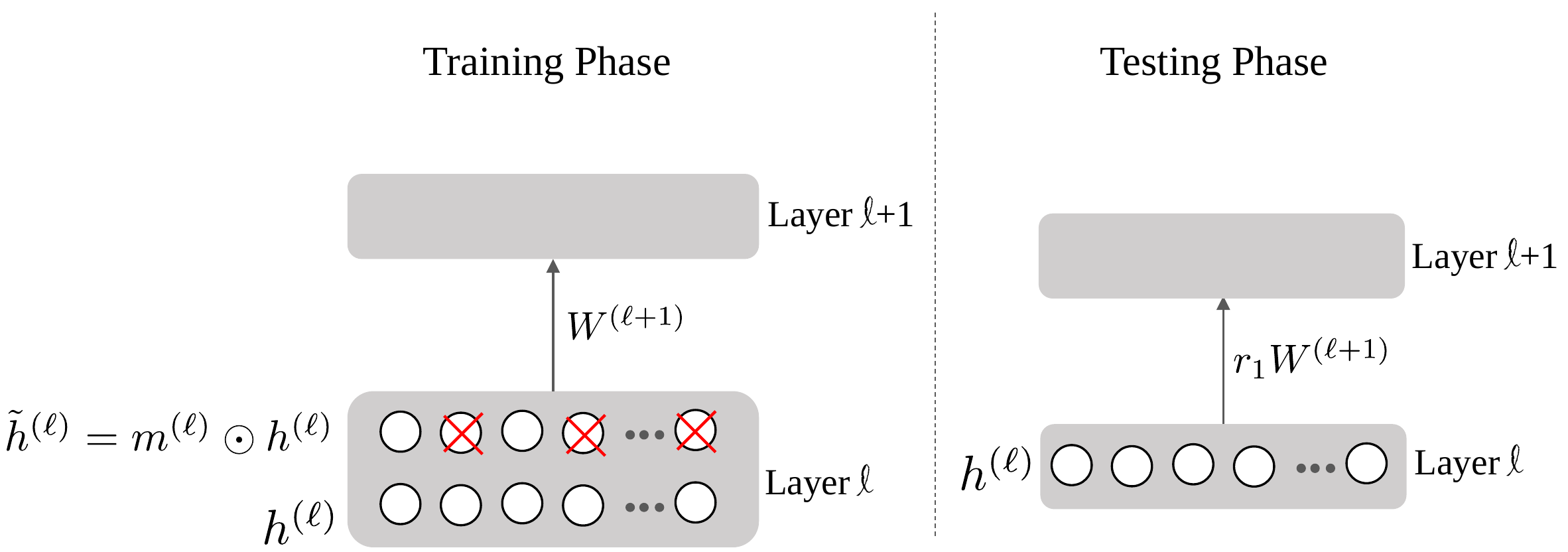}
\par\end{centering}

\protect\caption{Illustration of dropout in layer $\ell$ of a multilayer deep net.
\label{fig:Illustration-of-dropout-2}}
\end{figure}

In general, a neural network with more than one hidden layer can be
trained using a dropout rate $r_{1}$ for every layer. During testing,
all hidden units are retained and their outgoing weights are scaled
by a factor of $r_{1}$ (see Fig.~\ref{fig:Illustration-of-dropout-2}).
We describe the modified feed-forward and backpropagation equations
using dropout technique as follows. Consider a neural network with
$L$ hidden layers. For each layer $\ell$, where $\ell\in(1,2,\cdots,\,L)$,
let $h^{(\ell)}$ denote the hypothesis output, $z^{(\ell)}$ denote
the input to the layer and $b^{(\ell)}$ denote the bias. If the dropout
rate for the hidden layer is $r_{1}$, we generate $m^{(\ell)}$ --
a vector of independent Bernoulli random variables where each element
is $1$ with a probability $r_{1}$ and $0$ with a probability $(1-r_{1})$.
Hidden units in layer $\ell$ are dropped by element-wise multiplication
of $z^{(\ell)}$ and $m^{(\ell)}$. The modified feed-forward step
becomes:

\begin{align*}
m^{(l)} & =\text{Bernoulli}(r_{1})\\
\tilde{h}^{(\ell)} & =m^{(\ell)}\varodot h^{(\ell)}\\
z^{(l+1)} & =W^{(l+1)}\tilde{h}^{(\ell)}+b^{(l+1)}\\
h^{(l+1)} & =f\,(z^{(l+1)})
\end{align*}
where $f\,(z^{(l+1)})$ is the activation function of the hidden unit.

\subsection{Experiment settings for randomized methods \label{sub:Model-experiment-settings}}

The randomized methods are: Random Forests (RF), Gradient Boosting
Machine (GBM), and Deep Neural Networks with Dropout (DNND). Let $n$
be the size of training data, $p$ be the number of features, the
settings are as follows:
\begin{itemize}
\item \textbf{RF}: Number of trees is set at 25. Number of features per
split is $\sqrt{p}$, as often recommended in the RF literature. Leave
size is set at $\frac{n}{64}$, that is, there are maximally $64$
leaves per tree.
\item \textbf{GBM}: Number of weak learners is fixed at 200. Learning rate
$\lambda$ is not fixed for each learner, but starts from a small
value then increases until there is no improvement in the loss or
it reaches $0.1$. Data portion per weak learner is $\rho=0.5$, that
is, only 50\% of training data is used to train a weak learner. Each
weak learner uses a random feature subset of size $m=\min\left(\frac{p}{3},\sqrt{n}\right)$.
We use regression tree as weak learner, where the leave size is limited
to $\frac{n}{64}$. Following RF, at each node split, only a random
subset of features of size $\frac{m}{3}$ is considered.
\item \textbf{DNND}: We use a network with 2 hidden layers, 50 units each.
Although network sizes can be changed to fit the feature complexity,
we use the same architecture for all experiments to test its robustness.
Training is based on stochastic gradient descent in that parameter
is updated after every mini-batch of size $64$. Learning rate starts
at $0.1$ and is halved when the loss stops improving. Learning stops
when the learning rate falls below $10^{-4}$. Momentum of $0.9$
is used, and it appears to speed up the training. Regularization is
critical. We use three regularization methods: (i) Weight decay of
$10^{-4}$, which is equivalent to placing a Gaussian prior on the
weight; (ii) Max-norm of $1$ for weights combing to a hidden unit.
If the norm is beyond the prespecified max-value, the entire weight
vector is rescaled; (iii) Dropout rate of $0.5$ for both hidden units
and features. Applying dropout at feature level is critical to combat
against redundancy.\end{itemize}

\section*{References}


\end{document}